\documentclass[10pt,twocolumn,letterpaper]{article}

\usepackage[pagenumbers]{cvpr} 

\usepackage{graphicx}
\usepackage{amsmath}
\usepackage{amssymb}
\usepackage{booktabs}
\usepackage{enumitem}
\usepackage{multirow}

\usepackage[pagebackref,breaklinks,colorlinks]{hyperref}

\usepackage[capitalize]{cleveref}
\crefname{section}{Sec.}{Secs.}
\Crefname{section}{Section}{Sections}
\Crefname{table}{Table}{Tables}
\crefname{table}{Tab.}{Tabs.}
\usepackage{fancyvrb}
\usepackage{lipsum}
\makeatletter
\newcommand{\verbatimfont}[1]{\renewcommand{\verbatim@font}{\ttfamily#1}}
\makeatother

\usepackage{diagbox}

\makeatletter
\DeclareRobustCommand\onedot{\futurelet\@let@token\@onedot}
\def\@onedot{\ifx\@let@token.\else.\null\fi\xspace}

\def\eg{\emph{e.g}\onedot} 
\def\ie{\emph{i.e}\onedot} 
 
\def\etc{\emph{etc}\onedot} 
 
\def\etal{\emph{et al}\onedot}

\makeatother

\def\cvprPaperID{*****} 
\def\confName{CVPR}
\def\confYear{2022}

\begin{document}

\title{E$^2$TAD: An Energy-Efficient Tracking-based Action Detector}

\author{Xin Hu$^{2}$\thanks{The first three authors contribute equally to this paper. This work is done during Xin Hu and Hao-Yu Miao's remote internship at VITA Lab and Wormpex AI Research.},
Zhenyu Wu$^{2,4*}$\thanks{Project leader, email: $\langle \text{wuzhenyusjtu@gmail.com} \rangle$},
Hao-Yu Miao$^{2*}$,
Siqi Fan$^{2}$,
Taiyu Long$^{3}$,
Zhenyu Hu$^{2}$,
Pengcheng Pi$^{2}$, \\
Yi Wu$^{4}$,
Zhou Ren$^{4}$,
Zhangyang Wang$^{1}$\thanks{Corresponding author, e-mail: $\langle \text{atlaswang@utexas.edu} \rangle$ },
Gang Hua$^{4}$. \\
$^1$UT Austin \hspace{1em}
$^2$Texas A\&M University \hspace{1em}
$^3$New York University \hspace{1em}
$^4$Wormpex AI Research
}
\maketitle

\begin{abstract}
    Video action detection (spatio-temporal action localization) is usually the starting point for human-centric intelligent analysis of videos nowadays. It has high practical impacts for many applications across robotics, security, healthcare, \etc. The two-stage paradigm of Faster R-CNN in object detection inspires a standard paradigm of video action detection, \ie, firstly generating person proposals and then classifying their actions. However, none of the existing solutions could provide fine-grained action detection to the ``who-when-where-what'' level. This paper presents a tracking-based solution to accurately and efficiently localize predefined key actions spatially (by predicting the associated target IDs and locations) and temporally (by predicting the time in exact frame indices). This solution won the first place in the UAV-Video Track of 2021 Low-Power Computer Vision Challenge (LPCVC). The code is available at \url{https://github.com/VITA-Group/21LPCV-UAV-Solution}.
\end{abstract}

\section{Introduction}
The 2021 Low-Power Computer Vision Challenge (LPCVC) UAV-Video Track~\cite{hu20212020,hu2021e2vts} requires contestants to spatio-temporally localize the key action of ball-catching from videos captured by drones. The evaluation metric takes both accuracy and efficiency into account. The solution is expected to precisely detect the key action temporally (by predicting frame indices) and spatially (by predicting associated persons' and balls' IDs), using as low energy consumption as possible. This challenge has three unique challenges: lack of training data, robustness, and efficiency. 

\paragraph{Lack of Training Data:} Unlike the ubiquitous ``person'' object found in benchmarks for detection, segmentation, pose estimation, or tracking tasks, the ``ball'' object could only be found in COCO's sports ball category, with large semantic domain gap.

\paragraph{Robustness:} The irregular moving pattern of the actors and the drone has made the tracking extremely difficult. The resulting occlusion and varying view angle has brought enormous detection and association errors.

\paragraph{Efficiency:} Detecting target persons/balls, extracting their ReID features, and localizing key action spatio-temporally are computationally intensive. Considering the limited computation power and memory capacity on Pi 3B+, running these modules on Pi 3B+ in real-time would be difficult. 

This paper presents our submitted solution, dubbed Energy-Efficient Tracking-based Action Detector (E$^2$TAD). We present the following:
\begin{itemize}[leftmargin=*]
    \item a tracking-based vision system to spatio-temporally localize key action from videos. It has three core components: ball-person detection, deep association, and action detection;
    \item a harmonization-aware image composition module to generate synthetic but realistic ball with homogeneous color on pedestrian datasets;
    \item two adaptive inference strategies and a cache-friendly pipeline to save energy cost;
    \item a shape-texture debiased training and a domain-invariant adversarial training to improve the robustness of detection and ReID feature;
    \item a morphology-based and learning-free action detector that only depends on bounding box trajectories to localize key actions spatio-temporally.
\end{itemize}
This solution won the \textbf{1st place} in the UAV-Video Track of 2021 Low-Power Computer Vision Challenge (LPCVC).
\section{Related Works}
\subsection{Video Action Detection}
Video action detection (a.k.a, spatio-temporal video action localization) requires localizing persons and recognizing their actions in space and time from video sequences. 

STEP~\cite{yang2019step} is a progressive learning framework that consists of spatial refinement and temporal extension, which iteratively refines the coarse cuboid proposals toward action locations and incorporates longer-range temporal information to improve action classification.
STAGE~\cite{tomei2019stage} learns relations between actors and objects by self-attention operations over a spatio-temporal graph representation of the video. 
%
LFB~\cite{wu2019long} introduces a long-term feature bank that stores long-term supportive memory extracted over the entire video.
SlowFast~\cite{feichtenhofer2019slowfast} has a slow pathway operating at low frame rates to capture spatial semantic information, and a lightweight fast pathway operating at high frame rates to capture rapid motion better.
Recently, Context-Aware RCNN~\cite{wu2020context} enlarges the resolution of small actor boxes by cropping and resizing instead of RoI-Pooling, also extracting scene context information aided by LFB. 
AIA~\cite{tang2020asynchronous} leverages an interaction aggregation structure and an asynchronous memory update algorithm to efficiently model very long-term interaction dynamics.
ACAR-Net~\cite{pan2020actor}
learns to reason high-order relations (aka actor-context-actor relations) with a actor-context feature bank that preserves spatial contexts. 
Collaborative memory~\cite{yang2021beyond} shares the long-range context among sampled clips in a computation lightweight and memory-efficient way.
\subsection{Multi-Object Tracking}
\paragraph{Detection:}
In deep learning era, object detector can be grouped into two genres: ``two-stage detector'' and ``one-stage detector''.
One-stage detector classify and localize objects in a single shot using dense sampling. Two-stage detector has a separate module to generate region proposals. 
Compared with one-stage detectors, two-stage detectors usually achieve better accuracy but lower speed. RCNN series (RCNN~\cite{girshick2014rich}, Fast-RCNN~\cite{girshick2015fast}, Faster-RCNN~\cite{ren2016faster}, R-FCN~\cite{dai2016r}, and Mask-RCNN~\cite{he2017mask}) are the most representative two-stage detectors.
YOLO~\cite{redmon2016you,redmon2017yolo9000,redmon2018yolov3,bochkovskiy2020yolov4,ge2021yolox}, SSD~\cite{liu2016ssd}, and RetinaNet~\cite{lin2017focal} are the most representative one-stage detectors. 
Recently, anchor-free one-stage detectors have gained popularity, including CenterNet~\cite{zhou2019objects}, FCOS~\cite{tian2019fcos}, etc. 
An object detector's neck, which is composed of serveral bottom-up and top-down paths, is adopted to collect feature maps from different stages (scales). The most representative necks include FPN~\cite{lin2017feature}, PAN~\cite{liu2018path}, BiFPN~\cite{tan2020efficientdet}, and NAS-FPN~\cite{ghiasi2019fpn}.
\paragraph{Re-Identification:}
A standard closed-world Re-ID system has three main components: feature representation learning, deep metric learning, and rank optimization. 
Feature representation learning includes person-level global feature~\cite{zheng2017person}, part-level local features~\cite{zhao2017deeply,yao2019deep,sun2018beyond}, and auxiliary information (e.g., semantic attributes, viewpoint, and domain) enhanced features~\cite{su2016deep,chen2019abd,yuan2020calibrated}. 
Deep metric learning aims to design loss function to guide the feature representation learning. Verification loss~\cite{varior2016siamese,deng2018image,li2014deepreid}, identity loss~\cite{zheng2017person}, and triplet loss~\cite{hermans2017defense,chen2017beyond,yuan2020defense} are the three widely-used loss function. 
Ranking optimization improves the retrieval performance in the testing stage. 
\paragraph{Tracking:}
As the standard approach in multi-object tracking (MOT), \textit{Tracking-by-detection}~\cite{bewley2016simple,wojke2017simple,wang2020towards,zhou2020tracking,zhang2021fairmot,zhang2021bytetrack} has four stages: detection, feature extraction/motion prediction, affinity, and association. 
Given the raw frames, a typical workflow starts with an object detector that returns bounding boxes, a feature extractor and a motion predictor that give appearance and motion cues, an affinity calculator that computes the confidence of two objects belonging to the same target, and an object associator that assigns IDs to detected objects (if not excluded). 
MOT has two modes~\cite{ciaparrone2020deep}: batch and online. Since exploiting global information often results in better tracking, batch tracking uses future information when assigning IDs for detected objects in a certain frame. In comparison, online tracking, often running in real-time speed, cannot fix past errors using future information.

\subsection{Efficient Neural Network}
\paragraph{Pruning:} Pruning explores the redundancy in the model weights by removing the uncritical weights. It could be done in a one-shot way or an iterative way. One-shot methods prune weights at once based on some importance metric and require finetuning as accuracy loss compensation. Iterative methods prune weights during optimization following a progressively increasing sparsity rate. From a granularity perspective, ordered in decreased sparsity rate, pruning could be done element-wise, channel-wise, column-wise, filter-wise, or layer-wise. Channel-wise pruners include Network Slimming~\cite{liu2017learning}, AMC~\cite{he2018amc}, and ADMM~\cite{zhang2018systematic}. Filter-wise pruners include FPGM~\cite{he2019filter}, magnitude-based $\ell_1$ pruner~\cite{li2016pruning}, AGP~\cite{zhu2017prune}, Network Trimming~\cite{hu2016network}, and Taylor FO~\cite{molchanov2019importance}. Layer-wise pruners include NetAdapt~\cite{yang2018netadapt}. AutoCompress~\cite{liu2020autocompress} supports combined filter and column pruning.
\paragraph{Quantization:}
Quantization refers to compressing models by reducing the number of bits required to represent weights or activations, which can reduce the computations and the inference time. From granularity perspective, quantization could be done in layer-wise, group-wise, channel-wise (the standard method used for quantizing convolutional kernels), or sub-channel-wise. Deep neural networks' major numerical representation bitwidth is $32$-bit float (FP$32$). Many research works have demonstrated that weights and activations can be represented using $8$-bit integers~\cite{jacob2018quantization,zhou2016dorefa} without significant loss in accuracy. Lower bit-widths quantization, such as $4$/$2$/$1$ bits~\cite{courbariaux2016binarized,zhou2017incremental}, is an active field of research.

\paragraph{Dynamic Neural Networks:}
As opposed to static network's static computation graph and parameters, dynamic network, whose primary advantage is efficiency, can adapt its structures or parameters to the input during inference. Dynamic network allocates computation resources on-demand during inference, by selectively activating components (e.g., channels, layers, or stages) conditioned on the input~\cite{huang2017multi,lin2017runtime, shazeer2017outrageously,yang2019condconv,chen2020dynamic}. For example, a dynamic network spends less computation on easy samples or less informative spatial areas/temporal locations of an input. 
For image~\cite{wang2020glance,li2017dynamic} or video-related~\cite{li20212d,wang2021adaptive} tasks, sample-wise, spatial-wise, or temporal-wise adaptive inference could be conducted by formulating the recognition or detection task as a sequential decision problem and allowing early exiting during inference. 
\paragraph{Algorithm and Hardware Codesign:}
Algorithm and hardware codesign enables deep neural network cross the performance wall when parallelism and data reuse are exhausted. Efficient data operation and model compression are the two widely approaches. Stochastic data operation(e.g., data mapping, matrix decomposition) increases hardware utilization and accelerates computation on hardware~\cite{chen2019eyeriss, deng2019tie}. Compressed architecture (\eg, quantization and pruning) on accelerators ultra-highly speeds up inference time with negligible accuracy loss~\cite{lee201721mw, hegde2018ucnn}. Ren~\etal~\cite{ren2019admm} proposed the first algorithm-hardware co-designing framework with a combination of weight pruning and quantization, to reduce performance overhead due to irregular sparsity. Wang~\etal~\cite{wang2020time} extended Roofline model into deep learning area and incorporated computational complexity and run-time into models, which made it possible to analyze code performance for deep learning applications systematically.
\section{Approach}

\subsection{System Design}
The proposed \textbf{E}nergy-\textbf{E}fficient \textbf{T}racking-based \textbf{A}ction \textbf{D}etector (E$^2$TAD) has three core components: ball-person detection, deep association, and action detection.

\paragraph{Ball-Person Detection:}
\begin{figure}[!t]
    \captionsetup[subfigure]{labelformat=empty,justification=centering}
    \centering
    \includegraphics[width=\linewidth]{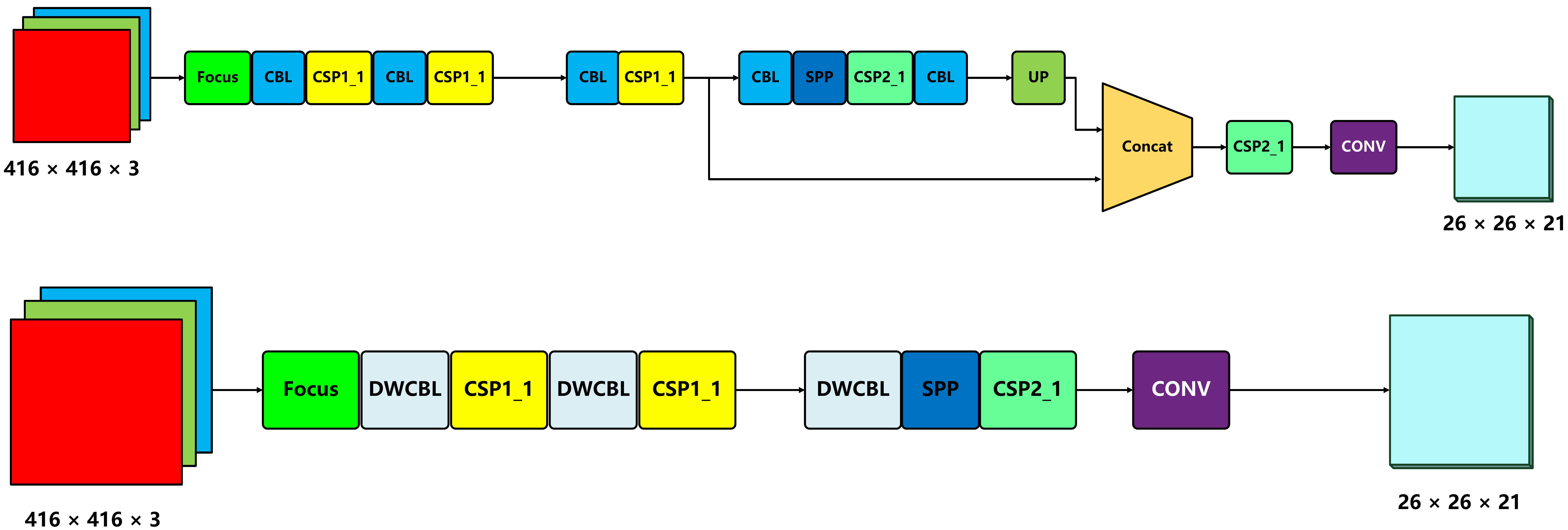}
    \caption{\small Proposed YOLO variants. The top figure shows YOLO-MobileV1 and the bottom figure shows YOLO-MobilbeV2.
    }
    \label{fig:yolo_mobile}
\end{figure}
Existing datasets are available for detecting pedestrians. However, there is no ball-related dataset in the literature. We adopts harmonization-aware image compositing~\cite{ling2021region} to generate synthetic but realistic balls on pedestrian datasets, \textit{e.g.}, VisDrone~\cite{cao2021visdrone}, COCO ball-person subset, PANDA~\cite{wang2020panda}.
To address the efficiency, we design a lightweight model based on YOLOv3 and YOLOv5. Among the three branches that detect objects in multi-scales (\textit{e.g.}, small, medium, and large), the middle one is retained to detect medium-scale objects. Our proposed two YOLO variants are dubbed YOLO-MobileV1 and YOLO-MobileV2.
Inherited from YOLOv5, YOLO-MobileV1 modifies PANet in YOLOv5 and cuts off one upsample module to directly output the medium-size feature map. 
As a descendant of YOLOv3, YOLO-MobileV2 replaces the convolution with depth-wise convolution and removes upsampling modules.
The architecture details of YOLO-MobileV1 and YOLO-MobileV2 are shown in Fig.~\ref{fig:yolo_mobile}.



\begin{figure}[!t]
    \centering
    \includegraphics[width=\linewidth]{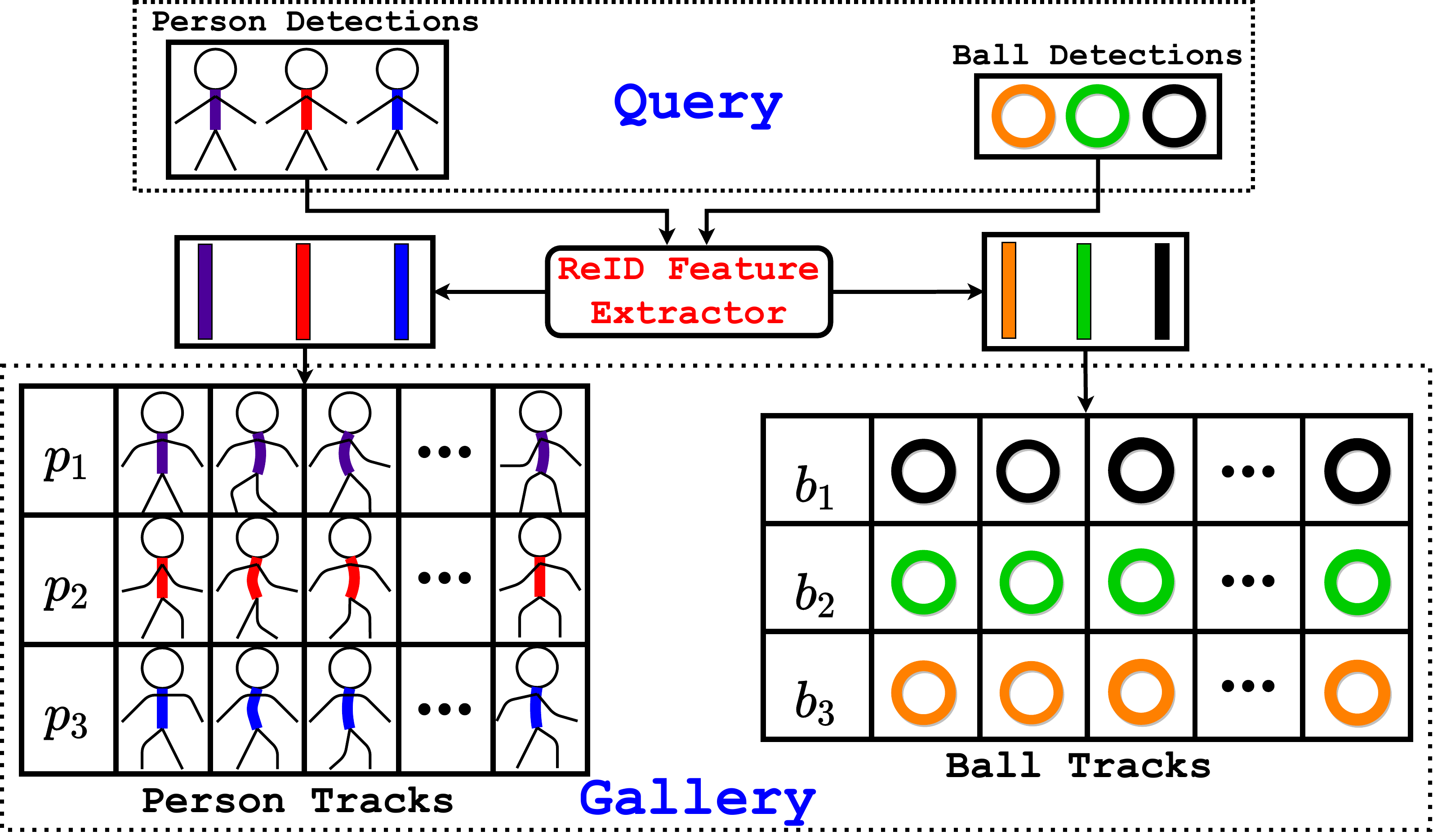}
    \caption{Proposed deep association that tags the detected persons and balls with their IDs.}
    \label{fig:deep_association}
\end{figure}

\paragraph{Deep Association:} Shown in Fig.~\ref{fig:deep_association}, this module tags the detected persons and balls with their corresponding identity labels. 
First, an identity-aware feature extractor (\textit{e.g.}, ResNet-18 trained on person ReID task) is adopted to obtain the feature embeddings of the detected persons/balls, given the bounding boxes together with the raw image. 
Second, Deep Association maintains the person tracklets and ball tracklets in a gallery. Each tracklet is a queue that stores ReID features of the latest $K$ tracks for the corresponding person/ball.
Last, using the ReID feature of newly detected persons/balls as queries and defining the cost of matching a query with a specific tracklet in the gallery as the distance between their ReID features, Deep Association formulates the identity tagging as a \textit{minimum cost assignment problem} that can be solved in polynomial time by Hungarian algorithm. To reduce association errors, any distance larger than a predefined threshold is marked as $\infty$ in the cost matrix to rule out the possibility of assignment.

\begin{figure*}[!t]
    \centering
    \includegraphics[width=\linewidth]{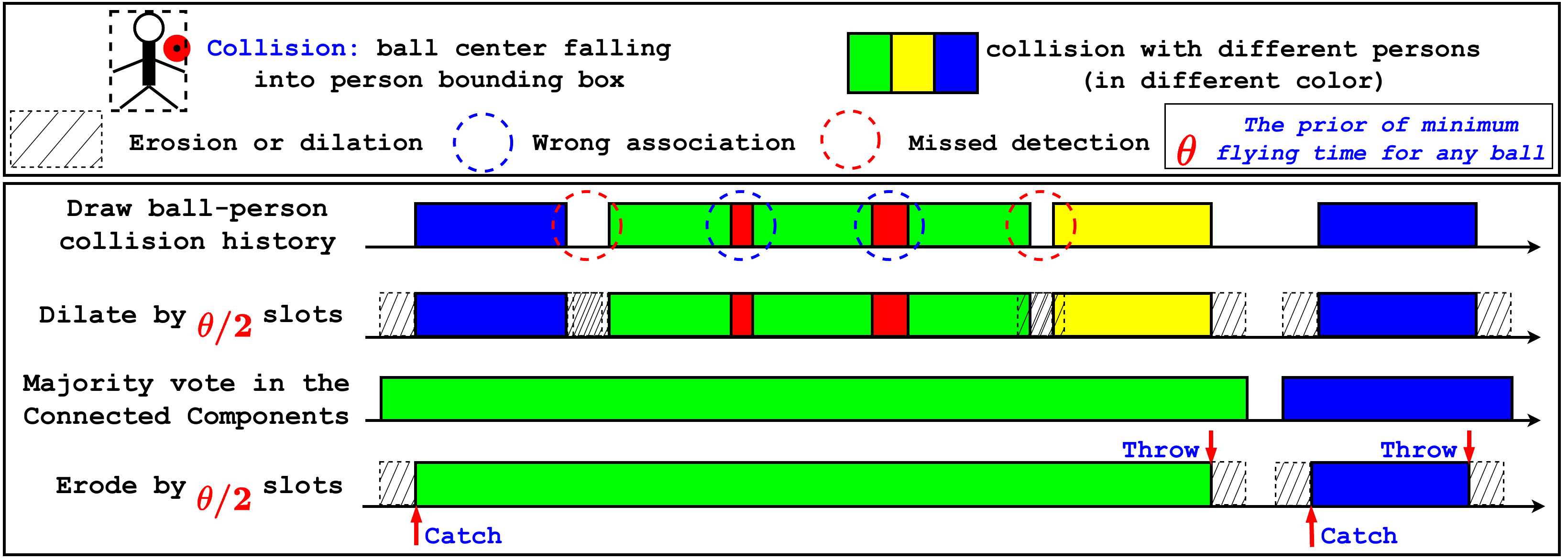}
    \vspace{-1.5em}
    \caption{\small Proposed heuristic approach for video action detection. The dilation/erosion parameter $\theta$ is the prior of minimum flying time for any ball.}
    \label{fig:action_detection}
\end{figure*}

\paragraph{Video Action Detection:} Free of learnable parameters, this module in Fig.~\ref{fig:action_detection} only depends on bounding box (bbox) trajectories to spatio-temporally localize key actions (\textit{e.g.}, catch and throw), thus extremely energy-efficient. A ball has two states: collision or non-collision. Collision happens when the center of one ball's bbox falls into one person's bbox. For each ball, the method plots its collision history with other persons as a gated signal spanning from $0$ to $T$, where $T$ is the total number of frames. The start and the end of any connected part in the gated signal signifies catch action and throw action, respectively.

\subsection{Efficiency}
\paragraph{Pruning:} Pruning could be done in an unstructured or structured way. Unstructured pruning removes individual weights at the kernel level, while structured pruning removes groups of weight connections such as channels or layers. Unstructured pruning leads to higher sparsity but requires specific hardware support for acceleration, thus not applicable to Raspberry Pi.

\paragraph{Quantization:} There are two types of quantization: \textit{post-training quantization} and \textit{quantization-aware training} (QAT)~\footnote{In QAT, a pre-trained model is quantized and then finetuned using training data to recover accuracy loss. In PTQ, a pre-trained model is quantized based on the clipping ranges and the scaling factors computed from the calibration data.}. 
Unlike \textit{post-training quantization}, QAT supports concurrent training and quantization. Consequently, such a simulation of quantization errors during training gives minimal accuracy loss if the representation bitwidth is converted from FP32 to INT8.

\begin{figure}[!t]
    \centering
    \includegraphics[width=\linewidth]{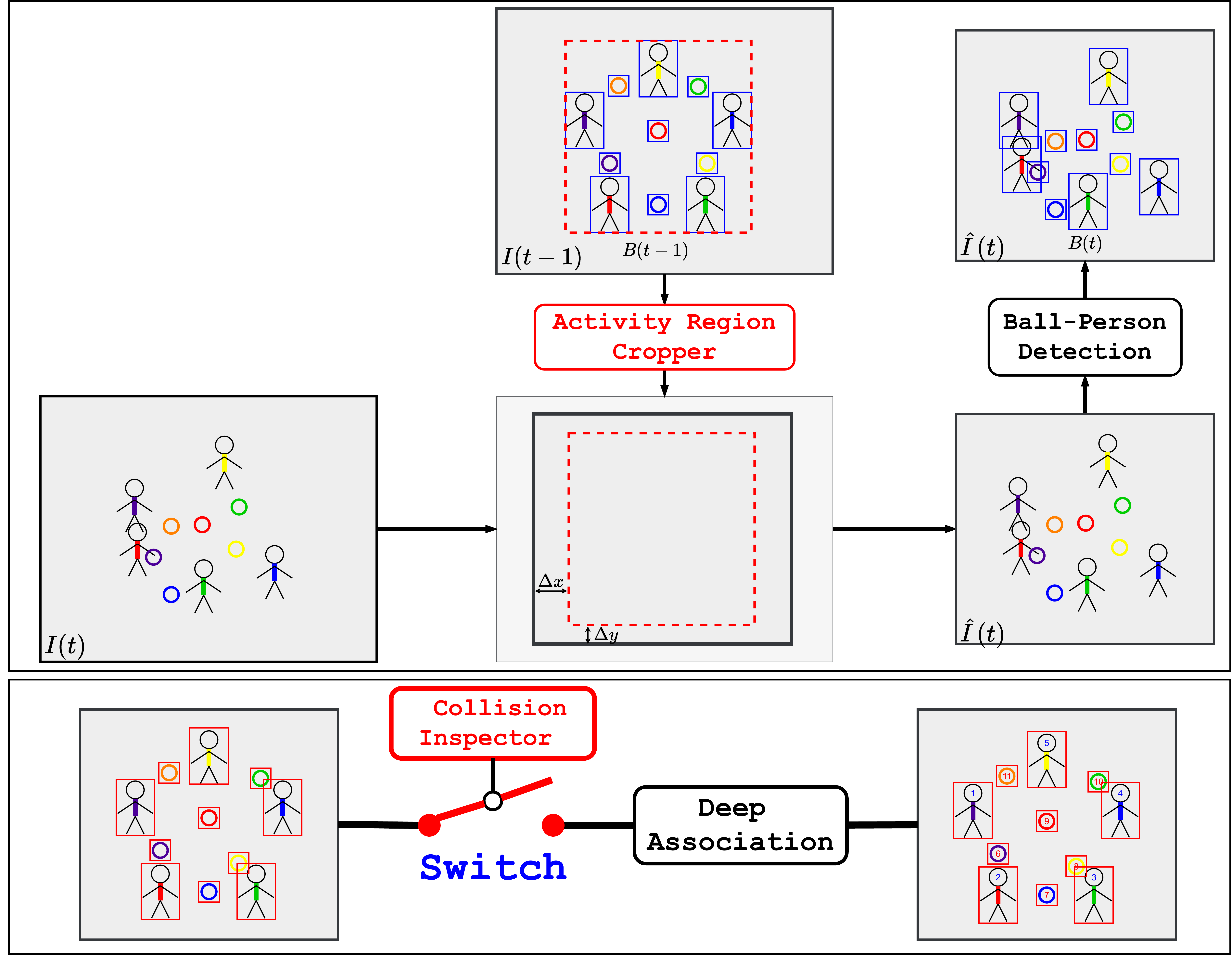}
    \caption{Proposed two data-dependent adaptive inference strategies: \textit{Activity Region Cropper} and \textit{Collision Inspector}.}
    \label{fig:adaptive_inference}
\end{figure}

\paragraph{Adaptive Inference:}
Considering video's temporal coherency and key action's infrequent occurrences, we are motivated to propose two adaptive inference strategies in Fig.~\ref{fig:adaptive_inference}: \textit{Activity Region Cropping} (ARC) and \textit{Collision Inspection} (CI).
Given the observation that persons and balls' locations not changing abruptly since videos are temporally coherent, ARC predicts the next frame's activity region ($\hat{I}_{t+1}$) based on the current frame's bboxes' coordinates ($B_t$) and crops out the non-activity region after adding some residual space ($\Delta x$ and $\Delta y$), whose details are shown below:
\begin{align*}
    \hat{I}_{t+1} = I_{t+1}[x_l-\Delta x:x_r+\Delta x, y_p-\Delta y:y_b+\Delta y],
    \label{activity_region_cropping}
\end{align*}
with $x_l, x_r, y_p, y_b=\min B_{t}^x, \max B_{t}^x, \min B_{t}^y, \max B_{t}^y$. ARC also improves the signal-to-noise ratio (SNR) of the input image for detection.
Since ball-person collision is a necessary condition for throwing or catching a ball, CI skips the succeeding Deep Association and Action Detection modules if no ball-person collision is inspected. 

\paragraph{Cache-friendly Pipeline:}
\begin{figure*}[!t]
    \centering
    \includegraphics[width=\linewidth]{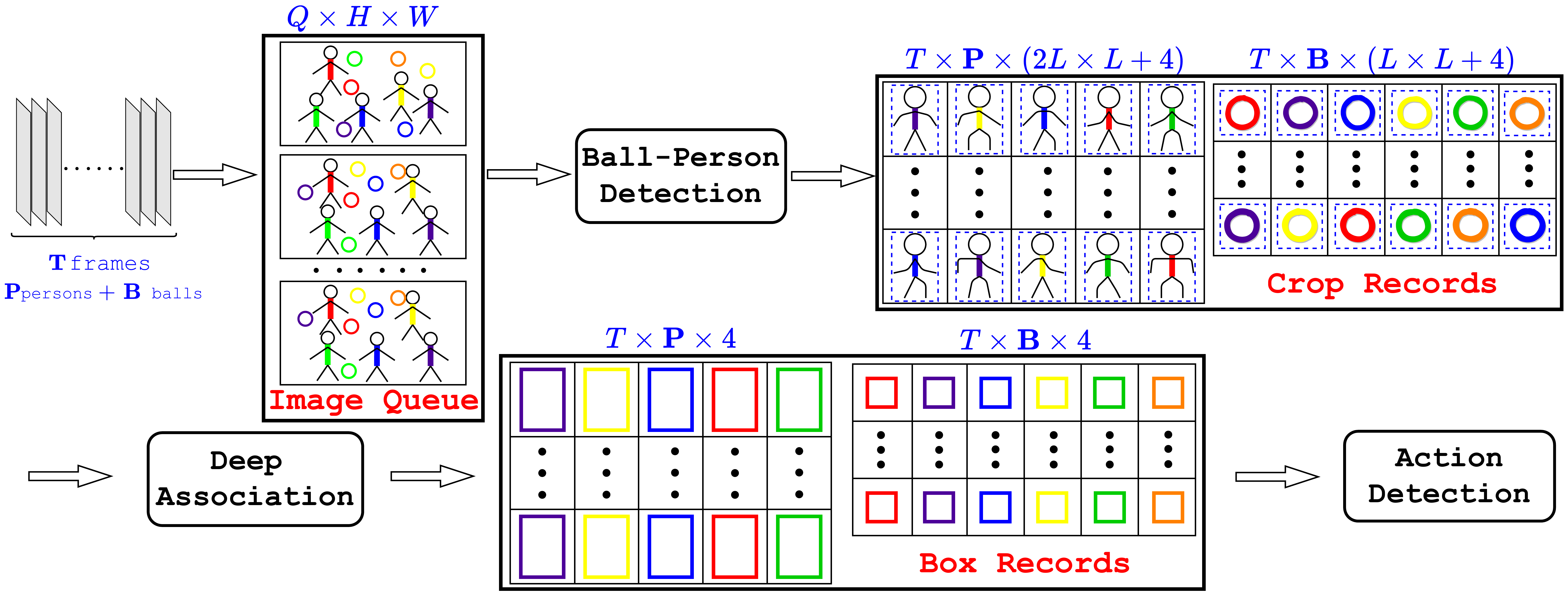}
    \caption{\small Proposed cache-friendly pipeline that trade memory for less data movement. Given a video of $T$ frames, which includes $\mathbf{P}$ unique persons and $\mathbf{B}$ unique balls, we store the intermediate decoded frames, person/ball crops, and person/ball boxes into the ``Image Queue'', ``Crop Records'', and ``Box Records'' respectively. The image queue size is $Q \times H \times W$ ($Q \ll T$). The person crop and the ball crop size are $2L \times L + 4$ and $L \times L + 4$ respectively. We set $H,W=540,960$ and $L=64$. The $4$ in crop size represents the $4$ box coordinates. Note that each box entry in ``Box Records'' is ordered by its corresponding person/ball ID.}
    \label{fig:optimal_pipelining}
\end{figure*}
There are three computation bottlenecks in the E$^2$TAD (ordered descendingly by profiling): video decoding, 
ReID feature extraction, and ball-person detection.
The computation pipeline of E$^2$TAD could be accelerated by addressing ``spatial locality'' and ``temporal locality'' for higher cache hit rate.
As is shown in the following code snippet and Fig.~\ref{fig:optimal_pipelining}, we store the intermediate results into an image queue ($Q \ll T$, where $Q$ as queue size and $T$ as the total number of frames), crop records ($T$ as the total number of frames and $H_c,W_c \ll H,W$), and bbox records ($N$ as the number of persons and balls), to reduce the \textit{Memory Access Cost} (MAC) for both model weights and data tensors. 
\vspace{+.5em}
\hrule
\vspace{-.5em}
\verbatimfont{\small}%
\begin{verbatim}
image_queue=zeros((Q,H,W,3),uint8)
crop_records=zeros((T,N,Hc+4,Wc+4,3),uint8)
bbox_records=zeros((T,N,4),int32)
\end{verbatim}
\vspace{-.5em}
\hrule
\vspace{+.5em}

To improve the temporal locality of ReID feature extraction, crop records and box records are used since the layer-wise computation could be fit in the L1 cache (32KB), and the weights after pruning and quantization could be fit in the L2 cache (512KB). To improve the spatial and temporal locality of video decoding, an image queue is used so that temporally consecutive frames could be decoded without interruption~\footnote{Video's temporal redundancy is utilized for high compression rate so that consecutive frames' encoded signals are closely packed together in storage.}.
Finally, E$^2$TAD's pipeline is turned from memory-access intensive to computation-intensive
Since the computation cost is fixed for a given video, the latency could be significantly reduced due to less memory access.
\subsection{Robustness}
\paragraph{Detection:}
Although shape and texture are two critical and complementary cues for object recognition, CNN-learned features are biased towards either shape or texture, depending on the training dataset. For example, the proposed YOLO-MobileV1 and V2 frequently mistake homogeneous color patches for balls. To alleviate the bias towards texture, we propose a \textit{texture-shape debiased training strategy} by compositing homogeneous color patches as hard negative samples on the training images. Moreover, we propose a geometry-aware circular anchor that takes the circle shape of balls into account to detect them better.

\begin{figure}[!t]
    \centering
    \includegraphics[width=\linewidth]{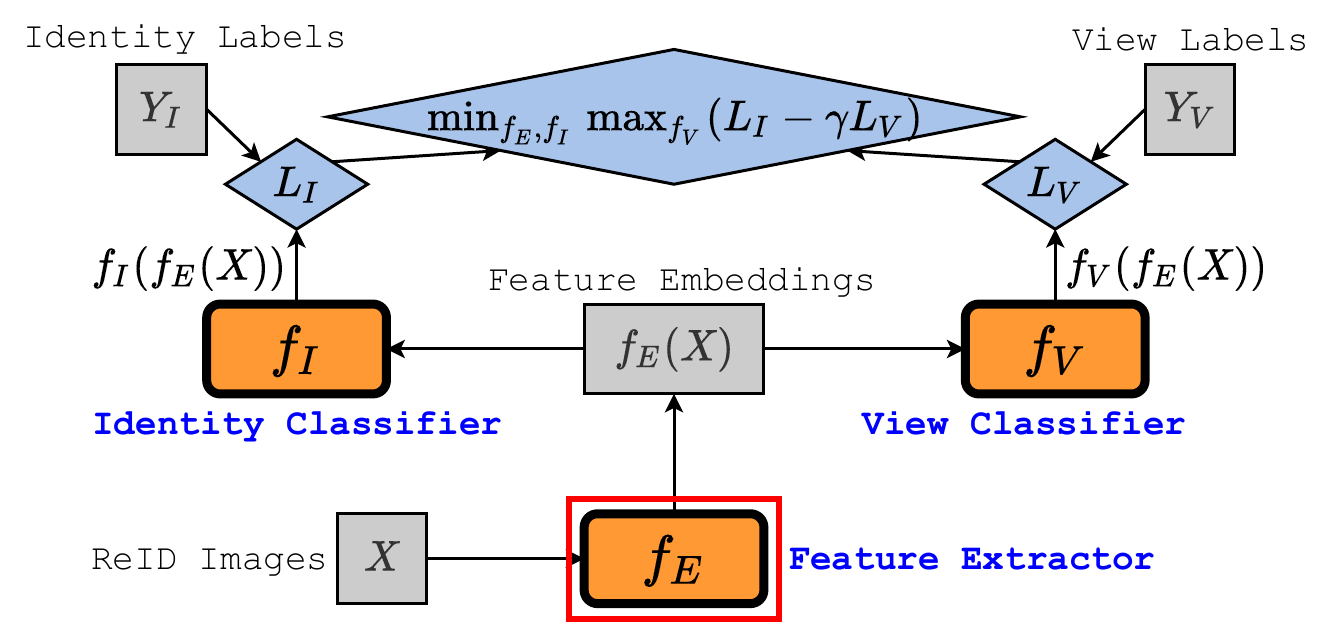}
    \caption{\small Proposed adversarial training in minimax game to learn robust ReID feature. After training, the identity classifier and the view classifier are discarded.}
    \label{fig:reid_minimax}
\end{figure}
\paragraph{ReID Feature:}
The two major challenges to robust ReID feature learning are various camera views and occlusion. Therefore, we are motivated to propose the following two solutions.
\textit{(1) Occlusion-Aware Data Augmentation:} Ball-person occlusion is mimicked by a circular binary mask overlayed on the person images.
We also use pseudo masks predicted by a pretrained person segmentation model to extract the person foreground and overlay it on the target person image to simulate person-person occlusion.
Both strategies are used as data augmentation to train the ReID model.
\textit{(2) Domain-Invariant Adversarial Feature Learning:}
Given training images $\mathcal{X}$, the goal is to learn a ReID feature embedding that is discriminative with different identities while invariant under different camera views. The goal can be mathematically expressed as a minimax game below~\cite{wu2018towards,wu2019delving,wu2020privacy,zheng2020pcal}:
\begin{equation*}
    \displaystyle\min_{f_e, f_{id}}\displaystyle\max_{f_v} L_{tr}(f_{id}(f_e(\mathcal{X})), \mathcal{Y}_{id}) - \lambda L_{ce}(f_v(f_e(\mathcal{X})), \mathcal{Y}_v).
    \label{minimax}
\end{equation*}
Here, the $f_e(\mathcal{X})$ is the extracted ReID feature embedding from $\mathcal{X}$. $f_{id}$ and $f_v$ predict identity labels and view labels from $f_e(\mathcal{X})$, respectively. $L_{tr}$ and $L_{ce}$ stands for triplet loss and cross-entropy loss. $\mathcal{Y}_{id}$ and $\mathcal{Y}_v$ symbolizes identity labels and view labels. $\lambda$ is a coefficient used to balance the two losses. More details are shown in Fig.~\ref{fig:reid_minimax}.

\paragraph{Action:}
As the downstream task, action detection is vulnerable to errors propagated from the upstream tasks (\textit{e.g.}, detection and association). In detection errors, false-negative cases may owe to occlusion, and false-positive cases could result from patches with homogeneous colors. Association errors arise from non-discriminative ReID features. Inspired by \textit{morphology-based denoising}~\cite{peters1995new}, we propose a heuristic approach in Fig.~\ref{fig:action_detection} to eliminate the preceding errors from the upstream tasks. As prior, $\theta$ is the minimum flying time for any ball. Given the collision history with different persons as a gated signal for each ball, we first dilate it by $\theta/2$. Then, we vote in the connected parts and select the majority as the associated person holding the ball, so that the preceding detection and association errors are eliminated. Last, the dilated gated signal is eroded by $\theta/2$.
\section{Experiments}
\subsection{Detection}
\paragraph{Datasets and Evaluation Protocols:} 
We evaluated the proposed ball-person detector on the sample videos provided by LPCVC21 UAV-Video Track, dubbed as \textit{LPCVC21}. Among the five sample videos, the two most challenging videos 7p3b and 5p5b are reserved for testing and dubbed as \textit{LPCVC21-test}; the remaining three videos 5p4b, 5p2b, and 4p1b are used for training and dubbed as \textit{LPCVC21-train}. The bounding boxes for persons are refined on the pseudo labels produced by a pretrained person detector and the bounding boxes for ball are manually labeled at every $15$ frames.
To increase the training set size, we mix PANDA~\cite{wang2020panda}, VisDrone pedestrian subset~\cite{cao2021visdrone}, and COCO ball-person subset~\cite{lin2014microsoft} with \textit{LPCVC21-train} to obtain a large-scale mixed dataset for ball-person detection, dubbed as \textit{MixDet-train}. \textit{MixDet-test} is the testing set of \textit{LPCV21}, namely \textit{LPCVC21-test}.

We adopt the COCO evaluation protocol's mAP$_{0.5}$ and mAP$_{0.5:0.95}$ to evaluate detection performance.

\begin{figure}[!t]
    \centering
    \includegraphics[width=\linewidth]{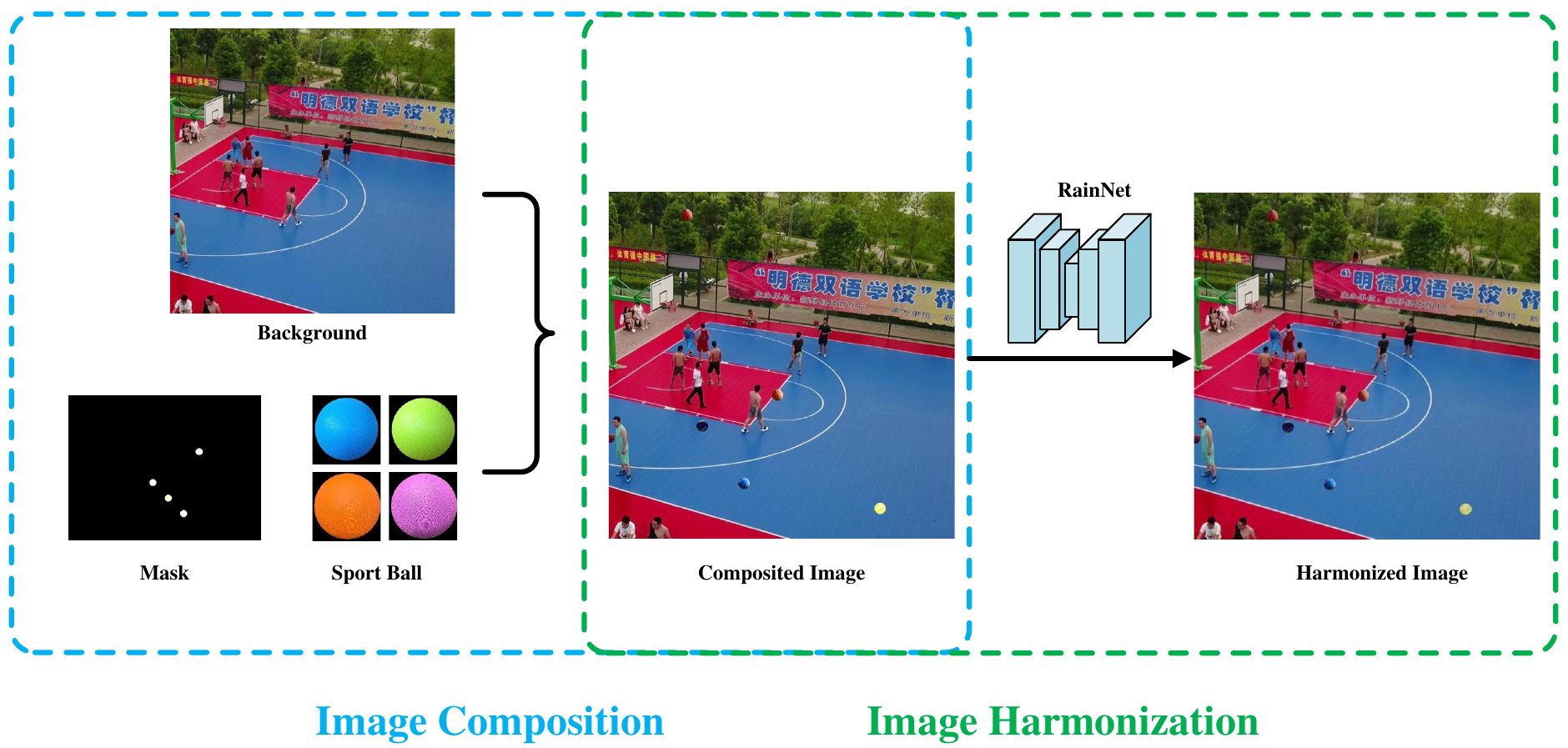}
    \caption{Overview of Harmonization-Aware Image Composition. Balls are composited into background images guided by randomly-placed masks. The composited images are harmonized by RainNet with region-aware instance normalization.}
    \label{fig:image_harmonization}
\end{figure}

\begin{figure}[!t]
    \captionsetup[subfigure]{labelformat=empty,justification=centering}
    \centering
    \subfloat
    {
        \includegraphics[width=0.48\linewidth]{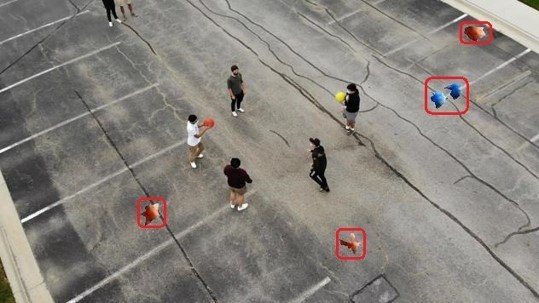}
    }
    \subfloat
    {
        \includegraphics[width=0.48\linewidth]{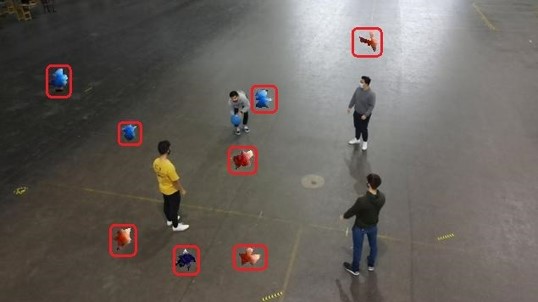}
    } 
    \caption{\small Shape-texture debiased training samples. Random patches with homogeneous color are marked by red rectangles.
    }
    \label{fig:debiased_training}
\end{figure}

\begin{figure}[!t]
    \captionsetup[subfigure]{labelformat=empty,justification=centering}
    \centering
    \subfloat
    {
        \includegraphics[width=0.48\linewidth]{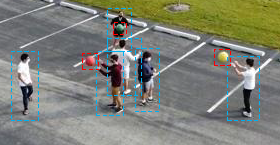}
    }
    \subfloat
    {
        \includegraphics[width=0.49\linewidth]{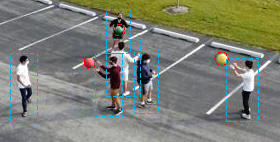}
    } 
    \caption{\small Samples of geometry-aware anchor for ``ball'' class. The left shows traditional anchors and the right shows geometry-aware anchor.
    }
    \label{fig:Geometry_anchor}
\end{figure}

\paragraph{Ablation Studies:} 
1. \textit{Harmonization-Aware Image Composition.}
Due to the lack of training data for the ball class, we propose a harmonization-aware image composition method to embed synthetic balls into images randomly selected from VisDrone and PANDA. Fig.~\ref{fig:image_harmonization} shows the two steps of harmonization-aware image composition. First, synthetic ball images are ``copy-pasted'' on background image guided by generated randomly-placed mask. Second, composited images are harmonized through RainNet~\cite{ling2021region} by taking the lighting and background into account.
2. \textit{Shape-Texture Debiased Training.}
To alleviate the ball-detection's bias towards texture, we add random-shaped patches with homogeneous texture, which serve as negative samples to detect balls. Debiased training images are showed in Fig.~\ref{fig:debiased_training}.
3. \textit{Geometry-Aware Anchor.}
Traditional rectangular bounding box (R-box) is not suitable for ball detection especially for heavily-occluded ball. The background information included in R-box introduces unexpected noise for ReID feature extraction. Considering the geometry difference between persons and balls, we use traditional rectangular bounding box (R-box) to detection persons and propose circular bounding box (C-box) to detect balls. Fig.~\ref{fig:Geometry_anchor} shows samples of different anchors. 

We conduct ablation studies of the proposed three methods in Tab.~\ref{table:detection_ablation}. Image-composition significantly improves detection performance. In addition, debiased training and circular anchor significantly improve both precision and recall, which shows that our proposed methods can effectively reduce false negatives (FN) and false positives (FP) for ball detection.

\begin{table}[h]
\centering
\caption{\small Ablation studies of proposed strategies to improve ball detection using YOLOv5-S.}
\resizebox{\linewidth}{!}{
\begin{tabular}{ccc|c|c|c|c}
\hline
ImgComp & DebiasTr & CircAnchor & Precision & Recall & $\text{AP}_{0.5}$ & $\text{AP}_{0.5:0.95}$  \\
\hline
- & - & - & 0.998 & 0.865 & 0.932 & 0.647  \\
\checkmark & - & - & 0.996 & 0.872 & 0.936 & 0.656  \\
\checkmark & \checkmark & - & 0.995 & 0.877 & 0.938 & 0.658  \\
\checkmark & \checkmark & \checkmark & 0.997 & 0.890 & 0.947 & 0.661  \\
\hline
\end{tabular}
}
\label{table:detection_ablation}
\end{table}
\paragraph{Results and Analyses:} We train YOLOv5s as baseline and the proposed YOLO variants on \textit{MixDet-train}, and test on \textit{MixDet-test}. Detection results are showed in Tab.~\ref{table:detection_comp}. YOLO-Mobile v1(416) reduces \textbf{16×} FLOPs at the cost of \textbf{24\%} mAP loss compared to YOLOv5-S(640). Furthermore, YOLO-Mobile v1(416) reduces \textbf{10×} energy compared to YOLOv5-S(640). We also propose three methods to advance detection performance.

\begin{table}[h]
\centering
\caption{\small Comparison of Different Detection Models.}
\resizebox{\linewidth}{!}{
\begin{tabular}{c|c|c|c|c|c|c}
\hline
Model & Input & FLOPs(G) & Params(M) & FPS & $\text{mAP}_{0.5:0.95}$ & Energy(kJ) \\
\hline
YOLOv5-S & 640 & 16.4 & 7.07 & 0.3 & 0.634 & 1.273 \\
YOLOv5-S & 416 & 6.9 & 7.07 & 0.7 & 0.575 & 0.579 \\
YOLO-MobileV1 & 640 & 2.7 & 1.08 & 5 & 0.535 & 0.242 \\
YOLO-MobileV1 & 416 & 1.2 & 1.08 & 10 & 0.485 & 0.142 \\
YOLO-MobileV2 & 416 & 0.8 & 0.241 & 14 & 0.356 & 0.146\\
\hline
\end{tabular}
}
\label{table:detection_comp}
\end{table}

\subsection{Re-Identification}
\paragraph{Datasets and Evaluation Protocols:}
We train and evaluate the ReID feature extractor on a mixed dataset from three person ReID datasets: CUHK03~\cite{li2014deepreid}, Market1501~\cite{zheng2015scalable}, and DukeMTMC~\cite{zheng2017unlabeled}, dubbed as \textit{MixReID}. \textit{MixReID-train} includes the training set of CUHK03/Market1501/DukeMTMC and the testing set of CUHK03/DukeMTMC. \textit{MixReID-test} is the testing set of Market1501.

For each query, we calculate the area under the Precision-Recall curve, known as average precision (AP). Then, the mean value of APs of all queries, i.e., mAP, is calculated.
\begin{table}[h]
\centering
\caption{\small Accuracy of different one-shot and iterative pruners under different sparsity rates.}
\resizebox{\linewidth}{!}{
\begin{tabular}{c|c|c|c|c|c}
\hline
\diagbox{Sparsity}{Pruner} & Baseline & $\ell_1$ & $\ell_2$ & FPGM & AGP \\
\hline
0   & 0.6550 & - & - & - & - \\
0.1 & - & 0.6468& 0.6389  & 0.6457& 0.6542\\
0.2 & - & 0.6451& 0.6321  & 0.6405& 0.6541\\
0.3 & - & 0.6412& 0.6257  & 0.6378& 0.6539\\
0.5 & - & 0.6385 & 0.6138 & 0.6316 & 0.6538\\
0.6 & - & 0.5517 & 0.5523 & 0.5730 & 0.5823\\
0.7 & - & 0.4781 & 0.4699 & 0.4926 & 0.5145\\
0.8 & - & 0.3948& 0.3896  & 0.4126& 0.4367\\
0.9 & - & 0.3395& 0.3287  & 0.3578& 0.3601\\
\hline
\end{tabular}
}
\label{table:reid_comp}
\end{table}

\begin{table}[h]
\centering
\caption{\small Tradeoff between accuracy and efficiency for AGP iterative pruner under different sparsity rates.}
\resizebox{\linewidth}{!}{
\begin{tabular}{c|c|c|c|c|c}
\hline
\diagbox{Metrics}{Sparsity} & 0 & 0.25 & 0.5 & 0.75 & 0.875 \\
\hline
FLOPS & 11.84 & 9.80 & 3.15 & 0.90 & 0.27 \\
Params & 11.16 & 7.94 & 2.79 & 2.00 & 0.17 \\
mAP & 0.6550 & 0.6540 & 0.6538 & 0.5830 & 0.4390\\
\hline
\end{tabular}
}
\label{table:reid_tradeoff}
\end{table}
\paragraph{Results and Analyses:}
In Tab.~\ref{table:reid_comp} and Tab.~\ref{table:reid_tradeoff}, ReID model is jointly scaled in its input resolution, depth, and width towards a desired efficiency-accuracy tradeoff. When scaling, structured pruning is adopted in an iterative (progressive) and adaptive (layerwise) way. FPGM~\cite{he2019filter} and AGP~\cite{zhu2017prune} pruner outperformed $\ell_1$ and $\ell_2$ filter pruner by achieving better efficiency-accuracy tradeoff.
Moreover, the input image resolution is reduced to $1/16$ (from $512 \times 256$ to $128 \times 64$)
\footnote{The input resolution for person crop is downsampled from $512 \times 256$ to $128 \times 64$ and ball crop is downsampled from $256 \times 256$ to $64 \times 64$.} 
and the ReID model is finetuned on the resized images.
Lastly, reducing bitwidth from FP32 to INT8 by QAT gives faster computation and lower memory usage for ReID: $3$ times inference time speed-up and $3.7$ times model size reduction with minor accuracy loss.

\subsection{Video Action Detection}
\paragraph{Datasets and Evaluation Protocols:}
We evaluate the video action detection performance on the two reserved videos for object detection testing, namely 7p3b and 5p5b in \textit{LPCVC21-test}. 
\begin{equation*}
    \label{eq:accuracy}    
    \text{Accuracy} = (\sum_{i=0}^{n} \frac{\text{correct}_i}{\text{total}_i})/(TP + 0.5\times(FP + FN))
\end{equation*}
The accuracy metric measure how well the action detector spatio-temporally predicts the ball-catching action. A True Positive (TP) predicts person and ball IDs correctly and the frame index within $\pm$10 frames as a tolerance threshold.
False positives and false negatives may arise from object detection and association errors.
\begin{table}[h]
\centering
\caption{\small Ablation studies of proposed strategies to improve robustness or efficiency. Score is computed as the ratio of accuracy and energy, \ie, Accuracy/Energy. $\uparrow$ stands for increased accuracy (by improving robustness) and $\downarrow$ indicates decreased energy cost (by improving efficiency).}
\resizebox{\linewidth}{!}{
\begin{tabular}{ccccc|c|c|c}
\hline
ReID$\uparrow$ & Det$\uparrow$ & Action$\uparrow$ & AdpInf$\downarrow$ & Pipeline$\downarrow$ & Accuracy & Energy & Score \\
\hline
 - & - & - & - & - & 0.657 & 0.261 & 2.517\\
\checkmark & - & - & - & - & 0.661 & 0.215 & 3.074\\
\checkmark & \checkmark & - & - & - & 0.734 & 0.135 & 5.437 \\
\checkmark & \checkmark & \checkmark & - & - & 0.771 & 0.135 & 5.711\\
\checkmark & \checkmark & \checkmark & \checkmark & - & 0.794 & 0.112 & 7.089 \\
\checkmark & \checkmark & \checkmark & \checkmark & \checkmark & 0.794 & 0.091 & 8.725\\
\hline
\end{tabular}
}
\label{table:action_detection}
\end{table}
\paragraph{Results and Analyses:}
ReID$\uparrow$ means robust ReID feature extractor improved by the proposed occlusion-aware data augmentation and domain-invariant adversarial feature learning. %
Det$\uparrow$ means robust ball detector improved by the proposed texture-shape debiased training.
Action$\uparrow$ means robust action detector improved by the proposed morphology-based denoising to eliminate preceding association and detection errors.
AdpInf$\downarrow$ means the two adaptive inference strategies implemented with an \textit{Activity Region Cropper} and a \textit{Collision Inspector}.
Pipeline$\downarrow$ means the cache-friendly pipeline that trades memory for less data movement cost.
Tab.~\ref{table:action_detection} presents the ablation study results, from which we can draw a conclusion that incrementally adding robust ReID features, robust ball detections, robust action detections, adaptive inference, and the cache-friendly pipeline could consistently improve the accuracy-efficiency tradeoff.

\begin{table}[h]
\centering
\caption{\small Performance of the top 4 winners' solution on the dry-run and test set. Our solution has a minimal score loss on the reserved test set. The robustness of E$^2$TAD accounts for the minimal score loss.}
\resizebox{0.95\linewidth}{!}{
\begin{tabular}{c|cccc|cc}
\hline
{} & \multicolumn{4}{c|}{Dry-run} & \multicolumn{2}{c}{Test} \\
\hline
Team & Rank & Energy & Accuracy & Score & Rank & Score \\
\hline
VITA & 1 & 0.091 & 0.790 & \textbf{8.569} & 1 & \textbf{8.473} \\
Meituan & 2 & 0.097 & 0.830 & 8.556 & 2 & 7.117 \\
ByteDance & 3 & 0.078 & 0.683 & 8.551 & 3 & 6.962 \\
ZJU+USYD+Beihang & 4 & 0.110 & 0.820 & 7.458 & 4 & 5.895 \\
\hline
\end{tabular}
}
\label{table:score_summary}
\end{table}

\paragraph{Performance on LPCVC's Dry-run and Test Set} Tab.~\ref{table:score_summary} gives the performance summary of the top 4 winners' solution on the dry-run and test set. Our proposed E$^2$TAD has a minimal score loss of $0.096$. The improved robustness of ball-person detector, ReID feature embedding, and action detector may account for it.
\section{Conclusion}
This paper is a tech report for our submitted solution to the UAV-Video Track of LPCVC-2021. Our proposed Energy-Efficient Tracking-based Action Detector (E$^2$TAD) is a tracking-based vision system that can spatio-temporally localize key action from videos. E$^2$TAD has three core components: ball-person detection, deep association, and action detection. First, we propose a harmonization-aware image composition module to generate synthetic but realistic balls. Second, we present two adaptive inference strategies, a cache-friendly pipeline, and some pruning and quantization strategies to address the energy-efficiency concern.
Third, we put forward a shape-texture debiased training and a domain-invariant adversarial training to improve the robustness of ball detection and ReID feature extraction. 
Last, inspired by morphology-based denoising, we develop a learning-free action detector that only depends on bounding box trajectories to spatio-temporally localize key actions.

\end{document}